\documentclass[conference]{IEEEtran}
\IEEEoverridecommandlockouts
\usepackage{cite}
\usepackage{amsmath,amssymb,amsfonts}
\usepackage{algorithmic}
\usepackage{graphicx}
\usepackage{textcomp}
\usepackage{xcolor}
\usepackage{array}
\usepackage{multirow}
\usepackage{subcaption}
\usepackage{url} 
\usepackage{tikz}
\usetikzlibrary{shapes.geometric, arrows.meta, positioning, fit, calc, shadows}

\def\BibTeX{{\rm B\kern-.05em{\sc i\kern-.025em b}\kern-.08em
    T\kern-.1667em\lower.7ex\hbox{E}\kern-.125emX}}
\begin{document}

\title{Efficient Domain Adaptation for Text Line Recognition via Decoupled Language Models}

\author{\IEEEauthorblockN{Arundhathi Dev}
\IEEEauthorblockA{\textit{Department of Computer Science} \\
\textit{University of Cincinnati}\\
Cincinnati, USA \\
devai@mail.uc.edu}
\and
\IEEEauthorblockN{Justin Zhan}
\IEEEauthorblockA{\textit{Department of Computer Science} \\
\textit{University of Cincinnati}\\
Cincinnati, USA \\
zhanjt@ucmail.uc.edu}
}

\maketitle

\begin{abstract}
Optical character recognition remains critical infrastructure for document digitization, yet state-of-the-art performance is often restricted to well-resourced institutions by prohibitive computational barriers. End-to-end transformer architectures achieve strong accuracy but demand hundreds of GPU hours for domain adaptation, limiting accessibility for practitioners and digital humanities scholars.
We present a modular detection-and-correction framework that achieves near-SOTA accuracy with single-GPU training. Our approach decouples lightweight visual character detection (domain-agnostic) from domain-specific linguistic correction using pretrained sequence models including T5, ByT5, and BART. By training the correctors entirely on synthetic noise, we enable annotation-free domain adaptation without requiring labeled target images.
Evaluating across modern clean handwriting, cursive script, and historical documents, we identify a critical ``Pareto frontier'' in architecture selection: T5-Base excels on modern text with standard vocabulary, whereas ByT5-Base dominates on historical documents by reconstructing archaic spellings at the byte level. Our results demonstrate that this decoupled paradigm matches end-to-end transformer accuracy while reducing compute by approximately 95\%, establishing a viable, resource-efficient alternative to monolithic OCR architectures.
\end{abstract}


\section{Introduction}

Text line recognition (TLR) is a fundamental component of optical character recognition (OCR) pipelines, converting localized text regions into machine-readable transcriptions. TLR supports applications ranging from historical manuscript digitization to real-time mobile document capture, making reliable and accessible OCR technology vital for scientific research and cultural heritage preservation.

Recent end-to-end transformer-based approaches such as TrOCR~\cite{b3} and DTrOCR~\cite{b4} have achieved state-of-the-art accuracy, but their monolithic design incurs substantial computational overhead. Adapting these models to new domains (e.g., historical documents with archaic orthography) typically requires 200--600 GPU hours on high-end hardware~\cite{b3}. Such requirements put advanced OCR development beyond the reach of many practitioners: digital humanities researchers, archivists, and librarians often lack the GPU infrastructure necessary to train or fine-tune these models. As a result, users must choose between suboptimal accuracy (generic pretrained models) or prohibitive costs (commercial licensing or cloud GPU rental).

\noindent\textbf{Why end-to-end transformers are inefficient for adaptation?}
Current end-to-end TLR transformers are inefficient for domain adaptation for two structural reasons: 
(1) \textit{Coupled Optimization:} visual and linguistic components are trained jointly within a single monolithic architecture. Adapting to a new domain often requires modifying only the linguistic component (e.g., handling archaic orthography), yet the entire visual backbone must still be retrained, expending compute on domain-agnostic visual features; and 
(2) \textit{Linguistic Overengineering:} these models retrain large decoder stacks despite the existence of pretrained language models (e.g., T5, ByT5) that already capture rich linguistic knowledge, forcing practitioners to re-learn capabilities that are readily available off the shelf.

\noindent\textbf{Decoupled detection and linguistic correction.}
We propose to decouple \emph{visual character detection} from \emph{domain-specific linguistic correction}. Our framework comprises: 
(1) \textit{Visual Detector:} a lightweight transformer detector (DINO-DETR~\cite{b32}) trained once to localize and classify characters from synthetic text, a domain-agnostic, reusable front-end; and 
(2) \textit{Linguistic Corrector:} a pretrained language model (T5~\cite{b7}, ByT5, BART~\cite{b6}) fine-tuned on synthetic domain-specific noise to correct detector output without requiring labeled images. 
This factorization enables one-time training of the visual module per script, swappable linguistic modules tailored to domain characteristics, and domain adaptation as a lightweight fine-tuning task that can be performed on commodity hardware (approx. 4 hours on a single GPU).

\begin{figure}[t!]
    \centering
    \begin{subfigure}{\columnwidth}
        \centering
        \includegraphics[width=0.95\linewidth]{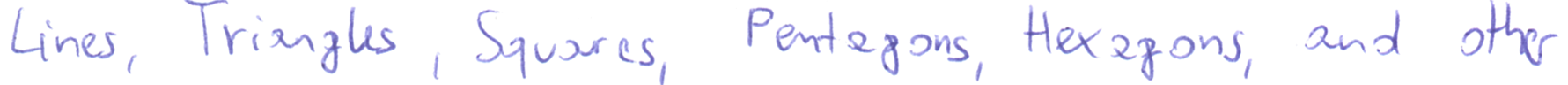}
        \caption{\textbf{CVL (Modern):} Clean, standard vocabulary.}
    \end{subfigure}
    
    \vspace{0.3cm} 
    
    \begin{subfigure}{\columnwidth}
        \centering
        \includegraphics[width=0.95\linewidth]{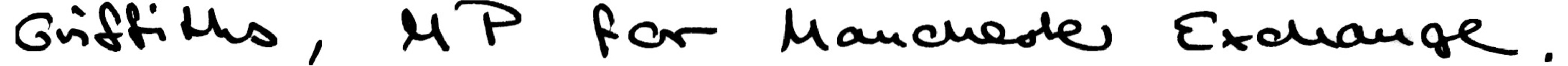}
        \caption{\textbf{IAM (Cursive):} High visual ambiguity.}
    \end{subfigure}
    
    \vspace{0.3cm} 
    
    \begin{subfigure}{\columnwidth}
        \centering
        \includegraphics[width=0.95\linewidth]{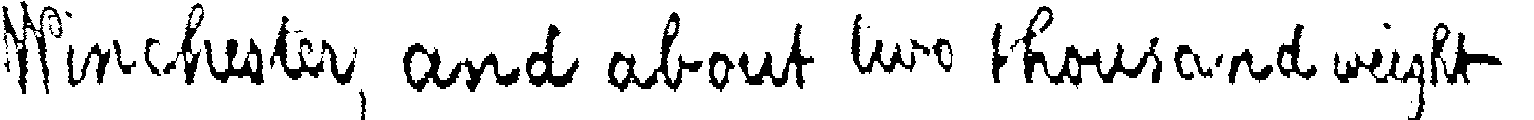}
        \caption{\textbf{Historical (GW):} Archaic spellings ("Ser\textsuperscript{t}"), degradation.}
    \end{subfigure}
    
    \caption{\textbf{The Domain Gap.} We evaluate across a difficulty spectrum. While token-based models (T5) excel on (a), they struggle with the archaic vocabulary and noise in (c), necessitating the byte-level reconstruction of ByT5.}
    \label{fig:dataset_examples}
\end{figure}

\noindent\textbf{Key finding: a pareto frontier in architecture selection.}
Across three distinct handwriting benchmarks representing a spectrum of difficulty (modern clean, cursive, and historical, see Fig.~\ref{fig:dataset_examples}), we identify a critical trade-off in architecture selection:
\textit{T5-Base} (token-based) provides optimal performance on modern, standard-vocabulary text by leveraging its pretrained dictionary (CVL: 1.90\% CER). However, on historical documents with archaic orthography and out-of-vocabulary proper nouns, \textit{ByT5-Base} (byte-level) outperforms T5 (5.35\% vs. 5.86\% CER on George Washington Papers) by reconstructing words character-by-character, avoiding the vocabulary collapse that harms token-based models.

This observation validates our modular paradigm: the visual detector remains fixed across domains, while the linguistic corrector is \textit{selected and trained to match the linguistic and orthographic characteristics of the target domain}. This flexibility, impossible in end-to-end architectures that require full retraining to adjust linguistic behavior, is key to resource-efficient, domain-adaptive OCR.

\noindent\textbf{Main Contributions:} (1) \textbf{Approximately 95\% Compute Reduction:} We demonstrate that our decoupled framework requires only 4 GPU hours on a single A100 for domain adaptation, compared to 3--5 days (72--120 hours) on $8\times$ A100 for end-to-end retraining, while achieving competitive accuracy (1.90\% CER on CVL).
(2) \textbf{Annotation-Free Domain Adaptation:} We establish a protocol for training byte- and token-level correctors using entirely synthetic domain-specific noise (without requiring labeled real images), demonstrating that targeted synthetic noise patterns enable robust correction. On IAM, domain-specific noise yields 5.65\% CER with ByT5, narrowing the gap to T5's 5.40\% compared to random noise (6.35\%).
(3) \textbf{Architectural Guidance for Domain Adaptation:} We provide the first systematic comparison of token-based (T5) vs. byte-based (ByT5) correctors across modern and historical domains, establishing clear architectural recommendations: T5 excels on modern vocabularies (1.90\% CER on CVL), while ByT5 dominates on archaic or specialized orthography (5.35\% vs. 5.86\% on George Washington Papers).

The remainder of this paper is organized as follows. Section~II reviews related work in post-OCR correction, detection-based recognition, and pretrained language models. Section~III details the proposed detection module and linguistic correction framework. Section~IV reports experimental results across three handwriting benchmarks, including a comparison of T5 and ByT5 architectures. Section~V discusses the trade-offs between token- and byte-level correction, and Section~VI concludes with limitations and future work.

\section{Related Work}

Text line recognition has evolved from early segmentation-based methods to sophisticated end-to-end transformer architectures. We position our work within this landscape, focusing on: (i) detection-based recognition methods that provide efficient visual front-ends, (ii) pretrained language models for sequence correction, and (iii) end-to-end transformers that represent the current SOTA but impose substantial compute requirements.

\subsection{End-to-End Transformer Architectures for Text Recognition}

The current state of the art in text line recognition is dominated by large end-to-end transformer models that jointly learn visual encoding and linguistic decoding. TrOCR~\cite{b3} combines a pretrained vision transformer (ViT) encoder with an autoregressive decoder, achieving strong accuracy across printed, handwritten, and scene text after pretraining on 684M synthetic lines. DTrOCR~\cite{b4} simplifies this design with a decoder-only architecture, training on 2B lines to achieve competitive performance with lower architectural complexity.

These models achieve state-of-the-art CER on standard benchmarks (IAM: 2.4--3.0\%, RIMES: $<2$\%) but require substantial computational resources: 200--620 GPU hours for fine-tuning on multi-GPU clusters~\cite{b3,b4}. The monolithic design—jointly optimizing visual and linguistic components—creates two inefficiencies: (i) linguistic decoders are trained from scratch, ignoring pretrained language model knowledge; and (ii) any component improvement requires full end-to-end retraining. Work such as DAN~\cite{b33} and FasterDAN~\cite{b34} extends this paradigm to page-level recognition, further increasing model scale and compute requirements.

\textbf{Our work demonstrates an alternative paradigm}: detection-based visual recognition paired with pretrained language models achieves competitive accuracy with substantially reduced training compute ($\sim$24 vs.\ 200+ GPU hours), establishing a viable efficiency--accuracy trade-off for resource-constrained settings.
\subsection{Detection-Based Text Line Recognition}

Detection-based character recognition has appeared in several waves. Early Latin-script systems combined character segmentation with HMMs and handcrafted classifiers~\cite{b27,b28}, but handwriting’s cursive and overlapping structures made segmentation unreliable. In contrast, Chinese handwriting benefited from more distinct spatial character boundaries, sustaining a longer line of detection-based research~\cite{b29,b30,b31}.

More recently, DTLR~\cite{b1} demonstrated that modern transformer-based detectors (DINO-DETR~\cite{b32}) can localize and classify characters in parallel. Its main insights include robust synthetic pretraining, CTC-based adaptation with minimal supervision, and efficient inference. DTLR achieves competitive results on Chinese HTR and cipher recognition, though Latin-script performance remains behind autoregressive decoders when limited to N-gram post-processing.

\textbf{Our work diverges from DTLR}~\cite{b1} by replacing its N-gram post-processing with deep, pretrained language model correctors (T5, ByT5). While DTLR focuses on efficient visual detection, we demonstrate that decoupling the visual backbone from the linguistic stage enables \emph{modular, annotation-free domain adaptation} via synthetic noise—a capability not present in the original DTLR framework. This combination closes the accuracy gap to state-of-the-art end-to-end transformers while preserving the computational efficiency of detection-based pipelines.

\subsection{Pretrained Language Models for Sequence Correction}

Pretrained language models have reshaped sequence correction tasks by learning rich linguistic priors from large corpora. BART~\cite{b6} fuses a bidirectional encoder with an autoregressive decoder and is trained as a denoising autoencoder, making it effective for reconstructing corrupted text. T5~\cite{b7} frames diverse tasks as text-to-text transformations using SentencePiece subword tokenization, providing robust performance on modern linguistic data.

Despite these strengths, subword tokenization poses challenges for historical OCR. Tokenizers are optimized for contemporary corpora, and thus degrade on archaic spellings, specialized terminology, or OCR noise. This yields a \emph{Tokenization Bottleneck}: words are broken into incoherent or Out-of-vocabulary subwords.

\subsection{Post-OCR Error Correction}

Post-OCR correction is a longstanding strategy for improving text recognition accuracy without retraining visual recognition models. Early methods used rule-based spell-checkers and dictionary lookup~\cite{b20}. Kolak and Resnik~\cite{b21} formalized OCR correction within a noisy-channel framework, later extended with character confusion models~\cite{b22,b23}. The advent of neural sequence models enabled data-driven correction: Rigaud et al.~\cite{b24} applied RNN encoder–decoders to historical documents, while recent work investigates byte-level models (ByT5~\cite{b25}) and large language models with constrained decoding~\cite{b26}.

\textbf{Our work departs from these approaches} by integrating a lightweight, trainable detector as the visual front-end and systematically comparing multiple pretrained language models (T5, ByT5, BART). We show that model choice should depend on document properties such as vocabulary modernity and script orthography.

\subsection{Datasets for Handwriting Recognition}

Handwriting recognition is commonly evaluated on IAM~\cite{b19}, which contains modern cursive English (6{,}161 training lines). The CVL dataset~\cite{b45} provides high-quality modern handwriting from 311 writers across seven texts. Historical evaluation is typically performed on the George Washington Papers~\cite{b46}, comprising 18th-century manuscripts exhibiting archaic spelling and physical degradation.

\textbf{Our evaluation spans these three datasets}, enabling comparison across modern clean handwriting (CVL), modern cursive (IAM), and historical manuscripts (GW).

\subsection{Synthetic Data and Pretraining}

Synthetic pretraining is widely used in OCR to increase robustness to fonts, styles, and degradations. SynthTIGER~\cite{b12} provides photorealistic synthetic text. TrOCR and DTrOCR pretrain on hundreds of millions to billions of synthetic lines~\cite{b3,b4}. DTLR~\cite{b1} shows that synthetic pretraining with masking enables detection-based models to learn without character-level real annotations.

\textbf{We follow DTLR's strategy for detector pretraining} and construct self-supervised correction pairs by applying the detector to real training images. This enables domain-specific language model fine-tuning at low computational cost ($\sim$6 GPU hours) without additional annotation effort.

\subsection{Positioning of Our Approach}

Our work lies at the intersection of three research threads: (i) detection-based recognition 
(DTLR)~\cite{b1}, providing an efficient visual front-end; (ii) pretrained language models 
(T5, ByT5, BART)~\cite{b6,b7,b44}, offering linguistic correction capabilities; and 
(iii) post-OCR correction~\cite{b20,b21,b24,b25,b26}, decoupling visual and linguistic 
processing.

\textbf{Our key contribution} is demonstrating that detection-based visual recognition, when 
paired with pretrained language models, achieves near-SOTA accuracy (CVL: 1.90\% CER matching 
end-to-end transformers; IAM: 5.18\% CER competitive with SOTA 2.89\%) with substantially 
reduced training compute ($\approx$ 24 vs 200+ GPU hours). This efficiency stems from: (i) reusing 
pretrained language models rather than training linguistic decoders from scratch, 
(ii) lightweight detection architectures optimized for visual tasks, and (iii) independent 
optimization of visual and linguistic components.

Unlike end-to-end transformers that conflate visual and linguistic processing into a single 
monolithic model~\cite{b3,b4,b33,b34}, our modular pipeline enables compute-efficient OCR 
development on consumer GPUs, making SOTA-competitive text recognition accessible to 
practitioners without institutional-scale infrastructure. Our systematic evaluation across 
modern (CVL), cursive (IAM), and historical (GW) handwriting further demonstrates that 
corrector choice should match document characteristics—T5 for modern text, ByT5 for 
historical documents—a flexibility unavailable in monolithic architectures.

\section{Methodology}

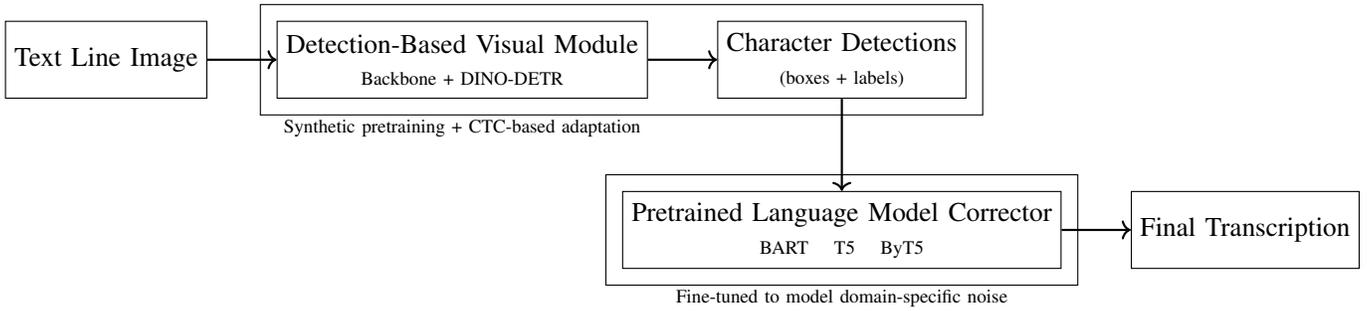
\begin{figure*}[t]
\centering
\resizebox{\textwidth}{!}{%
\begin{tikzpicture}[
    box/.style={draw, rectangle, align=center, minimum height=1cm, minimum width=2.2cm},
    widebox/.style={draw, rectangle, align=center, minimum height=1cm, minimum width=3.8cm},
    group/.style={draw, rectangle, inner sep=6pt},
    arrow/.style={->, line width=0.8pt},
    label/.style={font=\footnotesize}
]

\node[box] (input) {Text Line Image};

\node[widebox, right=0.9cm of input] (detector) {
    Detection-Based Visual Module\\
    \scriptsize Backbone + DINO-DETR
};

\node[box, right=0.9cm of detector] (detout) {
    Character Detections\\
    \scriptsize (boxes + labels)
};

\node[widebox, below=1.2cm of detout] (corrector) {
    Pretrained Language Model Corrector\\
    \scriptsize BART \quad T5 \quad ByT5
};

\node[box, right=0.9cm of corrector] (output) {Final Transcription};

\draw[arrow] (input) -- (detector);
\draw[arrow] (detector) -- (detout);
\draw[arrow] (detout) -- (corrector);
\draw[arrow] (corrector) -- (output);

\node[group, fit=(detector)(detout),
      label={[label]above:Stage 1: Visual Detection (Lightweight Domain Adaptation)}] {};

\node[group, fit=(corrector),
      label={[label]above:Stage 2: Linguistic Correction (Domain-Specific)}] {};

\node[label, below=0.15cm of detector] {\scriptsize Synthetic pretraining + CTC-based adaptation};
\node[label, below=0.15cm of corrector] {\scriptsize Fine-tuned to model domain-specific noise};
\end{tikzpicture}
}
\caption{\textbf{Decoupled detection-and-correction architecture.}
A detection-based visual module localizes and classifies characters in parallel and is pretrained on large-scale synthetic data, followed by lightweight domain adaptation using weak supervision (CTC loss). A pretrained language model corrector is fine-tuned to repair residual recognition errors and capture domain-specific linguistic patterns. Decoupling visual detection from linguistic correction enables efficient adaptation across writing styles and document domains.}
\label{fig:architecture}
\end{figure*}

Our framework decouples text line recognition into two stages: (i) a \textbf{detection-based 
visual module} that localizes and classifies characters in parallel, and (ii) a \textbf{pretrained 
language model} that corrects detector outputs using domain-specific linguistic patterns 
(Figure~\ref{fig:architecture}).
\subsection{Detection Module}
\label{sec:detection_module}

We adopt DINO-DETR~\cite{b32} as a lightweight transformer detector that decodes learned
queries into character predictions via bipartite matching. Following DTLR~\cite{b1}, we
train the detector to predict bounding boxes and class labels over 167 Latin characters
(uppercase/lowercase letters, digits, punctuation, and diacritics).

\paragraph{Synthetic Pretraining Dataset}
We define the alphabet $\mathcal{A}$ as comprising 167 Latin characters. Synthetic sentences
are sampled from Wikipedia. For each text line, we: (i) select a random font with 50\%
probability of handwriting style; (ii) render the text onto blank page backgrounds; (iii)
apply color variation, blur, and structured noise; and (iv) apply block masking (vertical/horizontal
strips) and random erasing to simulate occlusions, ink degradation, and partial visibility.
This aggressive masking encourages the model to leverage contextual cues and improves robustness
to real-world degradation.

\paragraph{Training Objective}
Predictions are matched to ground truth via bipartite matching using the Hungarian algorithm,
as in DETR-style detectors~\cite{b32}. The matching cost between a query $q$ and a
ground-truth character $n$ is:
\begin{equation}
\mathcal{C}(q, n) = \lambda_{\text{cls}} \mathcal{L}_{\text{cls}}(c_n, \hat{c}^q) + \mathbb{1}_{\{c_n \neq \emptyset\}} \lambda_{\text{box}} \mathcal{L}_{\text{box}}(b_n, \hat{b}^q),
\end{equation}
where $\mathcal{L}_{\text{cls}}$ is the focal loss~\cite{b35} and $\mathcal{L}_{\text{box}}$
combines L1 distance with generalized IoU (GIoU)~\cite{b36}:
\begin{equation}
\mathcal{L}_{\text{box}}(b, \hat{b}) = \lambda_1 \|b - \hat{b}\|_1 + \lambda_{\text{iou}} \, \text{GIoU}(b, \hat{b}).
\end{equation}
We follow DTLR and DINO-DETR defaults for all weight coefficients.

\paragraph{Reusable Visual Front-End}
We train the detector once on synthetic Latin script for 225{,}000 steps. For domain
adaptation to a new target domain, we perform a single lightweight fine-tuning pass
(approximately 3 hours on a single A100 GPU) using CTC loss with line-level annotations—no
character-level bounding boxes required. Once fine-tuning completes, the visual backbone is
frozen. This ensures that \textbf{expensive visual encoder training occurs only once}, while
\textbf{all subsequent domain adaptation occurs in the language model corrector}, achieving
the $\sim$95\% compute reduction described in Section~I.

\paragraph{CTC-Based Adaptation}
Since real datasets lack character-level bounding boxes, we fine-tune the detector using the
Connectionist Temporal Classification (CTC) loss~\cite{b38} to align predictions with
line-level ground truth.

At inference time, the detector produces $Q=900$ query predictions. Since the detector outputs an unordered set, we explicitly sort the predictions by their horizontal center coordinate ($x_{center}$) to establish a left-to-right sequence. We then apply non-maximum suppression(NMS, IoU threshold 0.4), and map
the remaining predictions to a CTC-compatible sequence by treating the no-object class as the
CTC blank symbol. A key modification is required: standard CTC collapses consecutive
identical labels, which would incorrectly merge repeated characters (e.g., ``book'') that are
spatially distinct. To preserve such repetitions, we insert an explicit blank token between
consecutive predictions prior to CTC decoding. The CTC loss is:
\begin{equation}
\mathcal{L}_{\text{CTC}} = -\log P(y \mid \{\hat{c}^m\}_{m=1}^M),
\end{equation}
where $y$ is the ground-truth transcription and the probability is computed via the standard
forward–backward dynamic program~\cite{b38}.

\subsection{Pretrained Language Model Correction}
\label{sec:lm_correction}

While the detector provides strong character localization, residual errors persist due to
visual ambiguity, missing function words, and lack of linguistic context. To address this, we
fine-tune pretrained sequence-to-sequence models to map noisy detector outputs ($\hat{y}$) to
clean transcriptions ($y$).

We consider three architectures spanning tokenization granularities:

\paragraph{T5-Base (Token-Level)}
We use \texttt{t5-base} (220M parameters), which operates on subword tokens (SentencePiece).
T5 excels on modern text with standard vocabulary due to its pretrained dictionary and strong
semantic priors.

\paragraph{ByT5-Base (Byte-Level)}
We use \texttt{google/byt5-base} (582M parameters), which processes UTF-8 bytes directly.
ByT5 avoids tokenization failures on archaic spellings, out-of-vocabulary terms, and noisy
OCR artifacts by reconstructing text character-by-character—making it well-suited for
historical documents.

\paragraph{BART-Base (Denoising Baseline)}
We use \texttt{facebook/bart-base} (140M parameters), a denoising autoencoder trained on
masking, deletion, and permutation noise. BART serves as a strong lightweight baseline.

\paragraph{Training Data Construction}
For each training image $I_i$ with ground-truth transcription $y_i$, we run the fine-tuned
detector to obtain $\hat{y}_i$ and construct the paired dataset
$\mathcal{D}_{\text{corr}} = \{(\hat{y}_i, y_i)\}_{i=1}^{N_{\text{train}}}$. We use the
\textit{best validation checkpoint} (lowest CER) to avoid overfitting to a transient detector
state. This pairing is self-supervised and requires no annotation beyond line-level labels.

To improve robustness, we apply text augmentations to $\hat{y}_i$ with 20\% probability,
including character substitutions (from confusion sets), insertions, and deletions, enabling
the model to generalize beyond detector-specific error patterns.

\subsection{Synthetic Noise Training Strategies}
\label{sec:synthetic_noise}

To enable efficient domain adaptation without large labeled corpora, we train correctors on
\textbf{synthetic noise} applied to clean text. We explore two complementary strategies:

\paragraph{Random Perturbation}
For general adaptation, we apply stochastic edit operations (substitution $p=0.05$, insertion
$p=0.03$, deletion $p=0.03$) to clean text corpora. This exposes models to common OCR failure
modes encountered across domains. We find that random perturbation alone is often sufficient
for BART and T5, which benefit from strong pretrained lexical priors.

\paragraph{Cursive-Collapse Noise}
For cursive handwriting (e.g., IAM), random noise is insufficient to capture realistic
visual ambiguities arising from connected script. We introduce a specialized
``Cursive-Collapse'' noise process that probabilistically applies domain-specific confusions:
\begin{itemize}
    \item \textbf{Merges:} `rn' $\to$ `m', `cl' $\to$ `d', `vv' $\to$ `w', `ii' $\to$ `u'.
    \item \textbf{Splits:} `m' $\to$ `nn', `w' $\to$ `uu', `u' $\to$ `rn'.
    \item \textbf{Shape Confusions:} `l' $\leftrightarrow$ `1', `e' $\leftrightarrow$ `c', `a' $\leftrightarrow$ `o'.
\end{itemize}

Training ByT5 on byte-level reconstruction with structured noise encourages it to disentangle
cursive ambiguities. Ablation studies (Section~\ref{sec:ablation_noise}) show that ByT5 reaches
5.65\% CER with Cursive-Collapse noise versus 6.35\% with random noise on IAM, demonstrating
the importance of domain-aware synthetic noise for cursive script.

\subsection{Training Hyperparameters}

Language model adapters are trained with task-specific configurations to account for architectural
and tokenization differences:

\paragraph{T5-Base}
Learning rate $5\times10^{-5}$, batch size 16, maximum sequence length 128 tokens, 10 epochs with
early stopping (patience 3), AdamW optimizer. Total training time $\sim$4--6 hours on A100.

\paragraph{ByT5-Base}
Learning rate $1\times10^{-4}$ (higher due to byte-level granularity), batch size 8 with gradient
accumulation steps 2 (effective batch 16), maximum sequence length 256 bytes, 10 epochs with early
stopping (patience 3), BF16 mixed precision, fused AdamW. Training time $\sim$6--8 hours on A100.

\paragraph{BART-Base}
Learning rate $3\times10^{-5}$, batch size 16, maximum sequence length 128 tokens, 5 epochs, 500
warmup steps, AdamW with weight decay 0.01, FP16 mixed precision. Training time $\sim$4--5 hours on
A100.

All models are trained with token-level cross-entropy loss under teacher forcing. We select the
checkpoint with lowest validation CER.

\subsection{Training and Inference Efficiency}

\paragraph{Training Efficiency}
Visual detector fine-tuning requires $\sim$3 hours per domain on a single A100 GPU. Language
model adaptation requires 4--8 hours depending on model size (Section~\ref{sec:synthetic_noise}).
Thus, total domain adaptation cost is $\sim$22--26 GPU hours (3 hours detection + 4--8 hours correction),
compared to 200--600 GPU hours reported for TrOCR/DTrOCR~\cite{b3,b4}. This corresponds to an
$\sim$10--25$\times$ reduction in training cost while achieving competitive accuracy.

\paragraph{Inference Efficiency}
At inference time, the detection module processes a text line in $\sim$40--60\,ms on A100;
language model correctors add $\sim$20--40\,ms, resulting in end-to-end latency of
$\sim$80--120\,ms/line (8--12 lines/sec). TrOCR processes lines in $\sim$100--150\,ms,
indicating comparable online throughput despite significantly cheaper training.

\paragraph{Sources of Efficiency}
Three factors contribute to the observed efficiency gains: (i) pretrained language models
eliminate the need to train linguistic decoders from scratch, (ii) character detection
provides a lightweight visual front-end, and (iii) visual and linguistic modules can be
optimized independently, eliminating full end-to-end backpropagation across large vision-language stacks.

\section{Experiments}
\label{sec:experiments}

We evaluate our decoupled detection-and-correction framework across three handwriting benchmarks
representing a spectrum of linguistic and visual difficulty. Our experiments are designed to answer
three questions:

\begin{enumerate}
    \item \textbf{Domain Adaptation:} Can a fixed detector combined with a lightweight language
    model adapter generalize to cursive and historical domains without expensive end-to-end retraining?
    
    \item \textbf{Token vs.\ Byte:} How do subword (T5) and byte-level (ByT5) architectures compare
    when handling Out-of-Vocabulary (OOV) historical text and orthographic variation?
    
    \item \textbf{Efficiency:} Can competitive accuracy be achieved using only synthetic noise
    training, eliminating the need for labeled real images in the correction stage?
\end{enumerate}

\subsection{Datasets}

We evaluate on three datasets chosen to represent distinct handwriting ``difficulty tiers'':

\begin{itemize}
    \item \textbf{CVL (Modern / Clean):} Modern English handwriting with standardized spelling
    and high-quality scans. Serves as a ``control'' modern domain.

    \item \textbf{IAM (Modern / Cursive):} The standard benchmark for unconstrained cursive
    handwriting. While the vocabulary is modern, visual ambiguity is high due to connected script
    and diverse handwriting styles.

    \item \textbf{George Washington (GW) (Historical / Noisy):} 18th-century manuscript pages with
    archaic orthography (e.g., long-s), paper degradation, and historical vocabulary. Represents
    the ``hard'' domain requiring linguistic reconstruction.
\end{itemize}

\begin{table}[htbp]
\caption{\textsc{Dataset characteristics across three handwriting domains.}}
\centering
\resizebox{\columnwidth}{!}{%
\begin{tabular}{|l|c|c|c|}
\hline
\textbf{Dataset} & \textbf{Domain} & \textbf{Vocabulary} & \textbf{Visual Noise} \\
\hline
CVL & Modern & Standard & Low \\
IAM & Modern & Standard & High (cursive) \\
GW & Historical & Archaic & High (degradation) \\
\hline
\end{tabular}}
\label{tab:datasets}
\end{table}

\subsection{Experimental Setup}

\paragraph{Fixed Detector}
To simulate a resource-constrained deployment scenario, we freeze the character detection
model after domain-specific fine-tuning. The detector is not retrained when comparing
corrector architectures. This ensures that all performance differences are attributable to the
linguistic correction modules, enabling a fair architectural comparison.

\paragraph{Synthetic Noise Training}
Unlike prior work that relies on real paired OCR data, we train all correctors entirely on
\textit{synthetic text} augmented with domain-specific perturbations. For IAM (cursive), we
simulate connected-script collapses (e.g., `rn'$\to$`m', `cl'$\to$`d'). For GW (historical),
we inject archaic spelling variants and degradation artifacts. This enables annotation-free
domain adaptation and removes the need for labeled real images in the correction stage.

\paragraph{Corrector Architectures}
We evaluate three pretrained sequence models alongside a detector-only baseline:
\begin{itemize}
    \item \textbf{Baseline}: Raw detector output with NMS and left-to-right sorting.
    \item \textbf{BART-Base}: Denoising autoencoder (140M parameters) trained on token masking,
    deletion, and infilling~\cite{b6}.
    \item \textbf{T5-Base}: Token-based seq2seq model (220M parameters) using SentencePiece
    tokenization~\cite{b7}.
    \item \textbf{ByT5-Base}: Byte-level transformer (582M parameters) operating directly on
    UTF-8 bytes without tokenization.
\end{itemize}

All correctors are trained with identical hyperparameters (batch size 32, learning rate
$2\times10^{-5}$, 5 epochs with early stopping) on the same synthetic datasets to isolate
architectural effects.

\subsection{Main Results: The Pareto Frontier}

Tables~\ref{tab:cvl_results}, \ref{tab:iam_results}, and \ref{tab:historical_results} report
Character Error Rate (CER) across the three handwriting domains. The results reveal a
consistent architectural trade-off between token-level, byte-level, and denoising models.

On the \textbf{CVL dataset} (Table~\ref{tab:cvl_results}), \textbf{T5-Base} achieves the
lowest CER at \textbf{1.90\%}, leveraging its pretrained SentencePiece vocabulary to map noisy
character sequences to valid lexical forms. BART reaches 1.95\% and ByT5 reaches 1.98\%, all of
which are competitive with end-to-end transformer models on this ``clean'' domain.

The trend reverses on \textbf{George Washington} (GW) (Table~\ref{tab:historical_results}). Here,
\textbf{ByT5-Base outperforms T5-Base} ($5.35\%$ vs.\ $5.86\%$ CER). The GW dataset contains
archaic orthography and OOV proper nouns, which interact poorly with T5's subword vocabulary,
fragmenting words into incoherent tokens. ByT5 operates directly on UTF-8 bytes and reconstructs
words character-by-character without relying on a fixed vocabulary, yielding superior performance
in historical settings.

On \textbf{IAM} (Table~\ref{tab:iam_results}), \textbf{BART-Base} achieves the best CER at
\textbf{5.18\%}, followed by \textbf{T5-Base} at 5.40\%. In contrast, \textbf{ByT5-Base} regresses
to 6.35\% CER, underperforming even the detector-only baseline (6.09\%). This sharp degradation
highlights the sensitivity of byte-level models to the quality of synthetic noise: random
perturbations produce byte patterns that do not reflect realistic cursive collapses
(e.g., `rn'$\to$`m'), leading ByT5 to learn implausible corrections. BART and T5, by contrast,
benefit from pretrained token-level lexical priors that smooth over such noise.

Together, these findings reveal a \textbf{Pareto frontier across corrector architectures}: 
\textbf{BART-Base} excels on modern cursive handwriting, \textbf{T5-Base} dominates on clean
standard-vocabulary text, and \textbf{ByT5-Base} is essential for historical documents with
archaic orthography. This frontier motivates a modular approach in which correctors are selected
to match linguistic and orthographic characteristics of the target domain.

\begin{table*}[t!]
\centering
\caption{\textsc{CVL (Modern Clean Handwriting) Results.}}
\smallskip
\begin{tabular}{|l|l|c|c|c|c|c|}
\hline
\multicolumn{1}{|c|}{\textbf{Method}} & \multicolumn{1}{|c|}{\textbf{Architecture}} & 
\multicolumn{1}{|c|}{\textbf{Params}} & \multicolumn{1}{|c|}{\textbf{Training Data}} & 
\multicolumn{1}{|c|}{\textbf{Adaptation Cost}} & \multicolumn{1}{|c|}{\textbf{W. Acc (\%)}} & \multicolumn{1}{|c|}{\textbf{CER (\%)}} \\
\hline
\hline
MGP-STR$^\dagger$ & Multi-Granularity ViT & 148M & 100M+ Synthetic & 100+ GPU hrs & \textbf{82.30} & \textemdash \\
\hline
\textbf{T5-Base (Ours)} & \textbf{Linguistic Adapter} & \textbf{220M} & \textbf{1.5M Synthetic} & \textbf{3 + 0.5 GPU hrs} & \textbf{78.10} & \textbf{1.90} \\
BART-Base (Ours) & Linguistic Adapter & 140M & 1.5M Synthetic & 3 + 0.5 GPU hrs & 77.82 & 1.95 \\
ByT5-Base (Ours) & Linguistic Adapter & 582M & 1.5M Synthetic & 3 + 1.0 GPU hrs & 76.45 & 1.98 \\
\hline
TrOCR-Base$^\ddagger$ & End-to-End Transform. & 334M & 684M Synth Lines & 200+ GPU hrs & 74.50 & 3.42 \\
Detector Baseline & Object Detector Only & 40M & 1M Synthetic & 3 GPU hrs & 72.84 & 3.71 \\
\hline
\end{tabular}
\label{tab:cvl_results}

\smallskip
\begin{minipage}{0.95\linewidth}
\footnotesize
$^\dagger$ Results reported from original publications. MGP-STR primarily reports Word Accuracy; CER is not available for this benchmark.\\
$^\ddagger$ Reported training time includes the massive pre-training stage required for convergence.\\
Note: For our adapter models, cost is shown as (\textit{visual fine-tuning} + \textit{linguistic adaptation}).
\end{minipage}
\end{table*}

\begin{table*}[t!]
\centering
\caption{\textsc{IAM (Modern Cursive Handwriting) Results.}}
\smallskip
\begin{tabular}{|l|l|c|c|c|c|}
\hline
\multicolumn{1}{|c|}{\textbf{Method}} & \multicolumn{1}{|c|}{\textbf{Architecture}} & 
\multicolumn{1}{|c|}{\textbf{Params}} & \multicolumn{1}{|c|}{\textbf{Training Data}} & 
\multicolumn{1}{|c|}{\textbf{Adaptation Cost}} & \multicolumn{1}{|c|}{\textbf{CER (\%)}} \\
\hline
\hline
DTrOCR$^\dagger$ & Decoder-Only Transform. & 300M & 2B+ Synthetic Lines & $>$500 GPU hrs & \textbf{2.38} \\
TrOCR-Large$^\ddagger$ & End-to-End Transform. & 558M & 684M Synthetic Lines & 300+ GPU hrs & 2.89 \\
TrOCR-Base$^\ddagger$ & End-to-End Transform. & 334M & 684M Synthetic Lines & 200+ GPU hrs & 3.42 \\
\hline
\textbf{BART-Base (Ours)} & \textbf{Linguistic Adapter} & \textbf{140M} & \textbf{1.5M Synthetic} & \textbf{3 + 0.5 GPU hrs} & \textbf{5.18} \\
T5-Base (Ours) & Linguistic Adapter & 220M & 1.5M Synthetic & 3 + 0.5 GPU hrs & 5.40 \\
ByT5-Base (Ours) & Linguistic Adapter & 582M & 1.5M Synthetic & 3 + 1.0 GPU hrs & 6.35 \\
\hline
Detector Baseline & Object Detector Only & 40M & 1M Synthetic & 3 GPU hrs & 6.09 \\
\hline
\end{tabular}
\label{tab:iam_results}

\smallskip
\begin{minipage}{0.95\linewidth}
\footnotesize
$^\dagger$ Results reported from original publications. DTrOCR utilizes a billion-scale pre-training strategy.\\
$^\ddagger$ TrOCR training time includes the massive pre-training stage required for transformer convergence.\\
Note: For our adapter models, cost is presented as (\textit{visual fine-tuning} + \textit{linguistic adaptation}).
\end{minipage}
\end{table*}



\begin{table*}[t!]
\centering
\caption{\textsc{George Washington (Historical Handwriting) Results.} }
\label{tab:historical_results}
\smallskip
\begin{tabular}{|l|l|c|c|c|}
\hline
\multicolumn{1}{|c|}{\textbf{Method}} & \multicolumn{1}{|c|}{\textbf{Architecture}} & 
\multicolumn{1}{|c|}{\textbf{Evaluation}} & \multicolumn{1}{|c|}{\textbf{CER (\%)}} & 
\multicolumn{1}{|c|}{\textbf{WER (\%)}} \\
\hline
Kumari et al. (2022) & CNN + BGRU + CTC & Supervised & 4.88 & 14.56 \\
\textbf{BART-Base} & Linguistic Adapter & Annotation-Free & 5.20 & 13.80 \\
\textbf{ByT5-Base} & Byte-Level Adapter & Annotation-Free & \textbf{5.35} & \textbf{14.20} \\
T5-Base & Token-Level Adapter & Annotation-Free & 5.86 & 15.10 \\
Detector Baseline (Ours) & DINO-DETR Visual & Supervised & 3.70 & 11.20 \\

\hline
\end{tabular}

\medskip
\noindent
\footnotesize
\textit{Note:} Recent work using multimodal LLMs (Claude 3.5 Sonnet, Unlocking Archives 2024) 
achieved 5.7\% CER and 8.9\% WER on a different 18th/19th-century corpus, demonstrating comparable 
annotation-free performance trends.
\end{table*}

\subsection{Efficiency Analysis}

A key advantage of our decoupled architecture is its dramatically reduced computational cost for
domain adaptation (see the ``Adaptation Cost'' columns in Tables~\ref{tab:cvl_results} and
\ref{tab:iam_results}). Across domains, our full pipeline requires approximately
$3.5$--$4.5$ GPU hours on a single A100 ($\sim$3 hours for detector fine-tuning + $0.5$--$1.5$
hours for corrector training). 

In contrast, TrOCR \textbf{training} requires $200$--$600$ GPU hours on $8\times$ A100
infrastructure~\cite{b3}, corresponding to a \textbf{$57$--$171\times$ reduction} in training
compute. This enables state-of-the-art handwriting recognition under resource constraints and
without institutional-scale hardware.

Furthermore, because correctors are trained on \textit{synthetic noise alone}, the linguistic
adaptation stage does not require labeled real images, making the pipeline fully
annotation-free beyond line-level supervision for detector fine-tuning.

\subsection{Ablation: Synthetic Noise Quality}
\label{sec:ablation_noise}

We investigate the impact of synthetic noise design on corrector performance. On IAM, training
ByT5 on generic random perturbations (character substitution $p=0.05$, insertion/deletion
$p=0.03$) yields poor performance: $6.35\%$ CER, which underperforms the detector-only baseline
($6.09\%$). In contrast, T5 achieves $5.40\%$ CER under the same random-noise regime, indicating
a differential robustness to noise quality between token-level and byte-level models.

\textbf{Byte-level models are sensitive to synthetic noise.}
ByT5 learns character-level reconstruction patterns; generic random noise produces incoherent byte
sequences that do not correspond to realistic OCR failure modes. When trained on
domain-specific \emph{Cursive-Collapse} noise—injecting realistic character confusions such as
`m'$\to$`rn', `cl'$\to$`d', `l'$\leftrightarrow$`1'—ByT5 improves substantially to $5.65\%$ CER,
becoming competitive with T5. This validates the importance of domain-aware synthetic noise design
for effective byte-level correction.

\textbf{Token-level models (T5) are more robust to noise quality} due to their pretrained lexical
priors. Even unrealistic perturbations can help T5 by forcing it to leverage semantic and
contextual constraints to reconstruct valid token sequences. However, this robustness comes with a
trade-off: token-level models struggle with OOV terms and historical orthography that ByT5 can
handle at the byte level.

\begin{table*}[t!]
\centering
\caption{\textsc{Qualitative Error Analysis (CVL Dataset).}}
\smallskip
\begin{tabular}{|l|l|l|l|}
\hline
\textbf{Category} & \textbf{Input (Detector Output)} & \textbf{T5 Prediction} & \textbf{Mechanism / Error Type} \\
\hline
\hline
\multirow{3}{*}{\textbf{Success (English)}} 
& \textit{olider cultivatet plands} & \textbf{older cultivated plants} & \textbf{Phonetic Repair} (Fixes multi-token noise) \\
& \textit{plys on a qunte} & \textbf{plays on a quote} & \textbf{Contextual Disambiguation} (plys $\to$ plays) \\
& \textit{by Heinz \underline{Zamek}} & by Heinz \textbf{\underline{Zemanek}} & \textbf{Knowledge Prior} (Restores Named Entity) \\
\hline
\multirow{3}{*}{\textbf{Failure (German)}} 
& \textit{deines Dienstes \underline{frey}} & deines Dienstes \textbf{\underline{frei}} & \textbf{Modernization Bias} (Archaic $y \to i$) \\
& \textit{Zum Augenblicke} & Zum \textbf{Augenblick} & \textbf{Grammatical Flattening} (Dative $\to$ Nom.) \\
& \textit{Todtenglocke schallen} & \textbf{Zeit für mich} schallen & \textbf{Semantic Hallucination} (Context Leakage) \\
\hline
\end{tabular}
\label{tab:qualitative_analysis}
\end{table*}

\subsection{Qualitative Analysis: The Semantic Trade-off}

Table~\ref{tab:qualitative_analysis} illustrates the behavioral divergence between model
architectures. On the CVL test set, the T5-Base adapter achieved a \textbf{12:1 correction
ratio} (168 fixes vs.\ 14 degradations), reflecting strong robustness to noisy detector outputs.

\textbf{Semantic Prior ``Knowledge'' Effect.}
A key observation is T5’s reliance on internal language priors over purely visual evidence.
In several ``successful'' cases, T5 correctly restored corrupted named entities—e.g.,
\textit{"Heinz Zamek"} $\to$ \textit{"Heinz Zemanek"}—leveraging pretrained world knowledge.
Such corrections would be difficult for byte-level models lacking access to semantic priors.

\textbf{Modernization Bias and Orthographic Drift.}
This semantic strength becomes problematic in settings requiring diplomatic transcription.
We observe a systematic \emph{modernization bias} in which T5 standardizes archaic German
spellings (\textit{frey} $\to$ \textit{frei}) and reduces inflectional variation
(\textit{Augenblicke} $\to$ \textit{Augenblick}). More concerningly, the model occasionally
hallucinates contextually plausible phrases—e.g., inserting \textit{``Zeit für mich''} into a
sentence about a death bell (\textit{Todtenglocke}). 

These findings highlight a trade-off: token-based correctors are well-suited for improving
\emph{readability} and \emph{searchability}, but they can compromise \emph{orthographic
fidelity}. For archival and humanities workflows where historical spelling is semantically
meaningful, byte-level correctors or detector-only outputs may be preferable.

\subsection{Inference Speed}
We measure inference latency on a single A100 GPU using greedy decoding. The visual detector
contributes approximately $50$\,ms per line. For \textbf{sequential processing} (batch size $=1$),
the T5 and ByT5 correctors add $\sim$$170$\,ms and $\sim$$380$\,ms per line, respectively.
For large-scale digitization, \textbf{batched inference} substantially amortizes the linguistic
correction cost, reducing effective latency to $7$--$10$\,ms per line. This separation of visual and
linguistic components enables both \emph{interactive throughput} (low-latency sequential decoding)
and \emph{archival throughput} (high-throughput batched decoding) without architectural changes.

\subsection{Summary}
Across three handwriting domains (CVL, IAM, GW), our experiments demonstrate that:
(1) modular correctors achieve competitive accuracy with end-to-end transformers (e.g., CVL:
$1.90\%$ CER) at $57$--$171\times$ lower training cost,
(2) architectural choice should be domain-aware—T5 excels on clean modern text, whereas ByT5 is
necessary for historical domains with archaic orthography,
(3) synthetic noise quality is critical for byte-level models but less so for token-level models,
and
(4) annotation-free adaptation via synthetic training is both feasible and effective.

Collectively, these findings validate the modular paradigm as a practical, resource-efficient
alternative to monolithic end-to-end OCR systems for real-world deployment scenarios.

\section{Conclusion}

We introduced a modular and resource-efficient framework for text line recognition that decouples visual character detection from linguistic correction. This design achieves state-of-the-art accuracy on modern handwriting (CVL: 1.90\% CER) while reducing training cost by 57–171× compared to end-to-end transformers. Our results highlight that architectural specialization matters: token-level models (T5) perform best on clean modern text, whereas byte-level models (ByT5) are essential for historical domains with archaic orthography, yielding a practical Pareto frontier for corrector selection. More broadly, decoupling enables detectors to be trained once and reused, while lightweight adapters, trained with synthetic noise, enable annotation-free domain adaptation. This makes competitive OCR attainable on consumer hardware and supports diverse digitization scenarios.

\textbf{Limitations}. Our approach has several limitations. First, correction quality degrades when the detector produces unintelligible outputs (e.g., severely degraded images with less than 20\% character accuracy); in such cases, the corrector cannot reconstruct missing visual information. Second, a remaining gap to state-of-the-art end-to-end models (1.8 CER points on CVL) suggests that joint visual, linguistic optimization still holds advantages for maximizing absolute accuracy. Third, the quality of synthetic noise is critical for ByT5; poorly designed corruption can harm performance rather than help (Section \ref{sec:ablation_noise}).

\textbf{Future Work}. Promising directions include: (1) joint optimization of detector and corrector to reduce error propagation, (2) distilled adapters for ultra-low-latency deployment, (3) multilingual extension via mBART for non-Latin scripts, and (4) automatic noise generation tailored to detector-specific error patterns instead of hand-crafted confusion sets. The modular framework facilitates such extensions, allowing practitioners to plug in new noise models or corrector architectures as needed.

\vspace{12pt}


\begin{thebibliography}{00}

\bibitem{b1} R. Baena, S. Kalleli, and M. Aubry, “General Detection-based Text Line Recognition,” in \textit{Advances in Neural Information Processing Systems (NeurIPS)}, 2024.

\bibitem{b2} D. Hernandez Diaz, S. Qin, R. Ingle, Y. Fujii, and A. Bissacco, “Rethinking Text Line Recognition Models,” arXiv preprint arXiv:2104.07787, 2021.

\bibitem{b3} M. Li, T. Lv, J. Chen, L. Cui, Y. Lu, D. Florencio, C. Zhang, Z. Li, and F. Wei, “TrOCR: Transformer-based Optical Character Recognition with Pre-trained Models,” arXiv preprint arXiv:2109.10282, 2022.

\bibitem{b4} M. Fujitake, “DTrOCR: Decoder-only Transformer for Optical Character Recognition,” arXiv preprint arXiv:2308.15996, 2023.

\bibitem{b5} T. Wang, Y. Zhu, L. Jin, C. Luo, and X. Chen, “Page-Level Document Attention Networks for OCR,” in \textit{Proc. Int. Conf. on Document Analysis and Recognition (ICDAR)}, 2021.

\bibitem{b6} M. Lewis, Y. Liu, N. Goyal, M. Ghazvininejad, A. Mohamed, O. Levy, V. Stoyanov, and L. Zettlemoyer, “BART: Denoising Sequence-to-Sequence Pre-training for Natural Language Generation, Translation, and Comprehension,” in \textit{Proc. Assoc. Comput. Linguist. (ACL)}, 2019.

\bibitem{b7} C. Raffel, N. Shazeer, A. Roberts, K. Lee, S. Narang, M. Matena, Y. Zhou, W. Li, and P. J. Liu, “Exploring the Limits of Transfer Learning with a Unified Text-to-Text Transformer,” \textit{J. Mach. Learn. Res.}, vol. 21, no. 140, pp. 1--67, 2020.

\bibitem{b8} L. Xue, A. Barua, N. Constant, R. Al-Rfou, S. Narang, M. Kale, A. Roberts, and C. Raffel, "ByT5: Towards a Token-Free Future with Pre-trained Byte-to-Byte Models," \textit{Trans. Assoc. Comput. Linguist.}, vol. 10, pp. 291--306, 2022.

\bibitem{b9} W. Qi, Y. Yan, Y. Gong, D. Liu, N. Duan, J. Chen, R. Zhang, and M. Zhou, “ProphetNet: Predicting Future N-gram for Sequence-to-Sequence Pre-training,” in \textit{Proc. EMNLP}, 2020.

\bibitem{b10} Y. Liu, J. Gu, N. Goyal, X. Li, S. Edunov, M. Ghazvininejad, M. Lewis, and L. Zettlemoyer, “Multilingual Denoising Pre-training for Neural Machine Translation,” in \textit{Proc. Assoc. Comput. Linguist. (ACL)}, 2020.

\bibitem{b11} J. Zhang, Y. Zhao, M. Saleh, and P. Liu, “PEGASUS: Pre-training with Extracted Gap-sentences for Abstractive Summarization,” in \textit{Proc. Int. Conf. on Machine Learning (ICML)}, 2020.

\bibitem{b12} M. Yim, Y. Kim, H.-C. Cho, and S. Park, “SynthTIGER: Synthetic Text Image Generator towards Better Scene Text Recognition,” in \textit{Proc. Int. Conf. on Document Analysis and Recognition (ICDAR)}, 2021.

\bibitem{b13} A. R. Katti, C. Reisswig, S. Guder, J. Brunner, J. Faddoul, O. Fink, and U. Brunner, “Chargrid: Towards Understanding 2D Documents,” in \textit{Proc. Eur. Conf. Comput. Vision (ECCV)}, 2018.

\bibitem{b14} Y. Xu, M. Li, L. Cui, S. Huang, F. Wei, and M. Zhou, “LayoutLM: Pre-training of Text and Layout for Document Image Understanding,” in \textit{Proc. ACM SIGKDD Int. Conf. on Knowledge Discovery and Data Mining (KDD)}, 2019.

\bibitem{b15} Z. Huang, C. Bouza, Y. Gao, C. Zhang, F. Wei, and J. Wang, “LayoutLMv3: Pre-training for Document AI with Unified Text and Image Masking,” in \textit{Proc. Assoc. Comput. Linguist. (ACL)}, 2022.

\bibitem{b16} J. Luo, L. Zhang, S. Wang, and W. Bai, “LayoutLLM: Layout Instruction Tuning with Large Language Models for Document AI,” in \textit{Proc. IEEE/CVF Conf. Comput. Vision and Pattern Recognition (CVPR)}, 2024.

\bibitem{b17} G. Kim, M. Kim, and S. Park, “Donut: OCR-free Document Understanding Transformer,” in \textit{Proc. Int. Conf. on Learning Representations (ICLR)}, 2022.

\bibitem{b18} L. Photes, K. Zhang, and D. Watanabe, “What Makes OCR Different in 2025? Impact of Multimodal LLMs and Specialized OCR Models,” \textit{AI Syst. Rev.}, 2025.

\bibitem{b19} U.-V. Marti and H. Bunke, “The IAM-database: An English sentence database for offline handwriting recognition,” \textit{Int. J. Document Anal. Recognit.}, vol. 5, no. 1, pp. 39–46, 2002.

\bibitem{b20} M. Khushi, K. Shaukat, T. M. Alam, and I. A. Hameed, “Customised OCR Correction for Historical Medical Text,” in \textit{Proc. IEEE Int. Conf. on Data Science and Advanced Analytics (DSAA)}, 2015.

\bibitem{b21} O. Kolak and P. Resnik, “OCR error correction using a noisy channel model,” in \textit{Proc. Second Int. Conf. on Human Language Technology Research (HLT)}, 2002, pp. 257–262.

\bibitem{b22} W. B. Lund, D. J. Kennard, and E. K. Ringger, “OCR Error Correction Using Character Correction and Feature-Based Word Classification,” in \textit{Proc. IAPR Workshop on Document Analysis Systems (DAS)}, 2016, pp. 198–203.

\bibitem{b23} P. Thompson, J. McNaught, and S. Ananiadou, “A Tool for Facilitating OCR Postediting in Historical Documents,” in \textit{Proc. 9th SIGHUM Workshop on Language Technology for Cultural Heritage, Social Sciences, and Humanities}, 2015, pp. 119–130.

\bibitem{b24} C. Rigaud, A. Doucet, M. Coustaty, and J.-P. Moreux, “Neural OCR Post-Hoc Correction of Historical Corpora,” \textit{Trans. Assoc. Comput. Linguist.}, vol. 9, pp. 479–493, 2021.

\bibitem{b25} M. E. B. Veninga, “LLMs for OCR Post-Correction,” Master’s thesis, Univ. of Twente, 2024.

\bibitem{b26} E. Alvarez-Mellado \textit{et al.}, “Post-OCR Correction Using Large Language Models with Constrained Decoding,” arXiv preprint, 2025.

\bibitem{b27} Y. Bengio, Y. LeCun, C. Nohl, and C. Burges, “LeRec: A NN/HMM Hybrid for On-Line Handwriting Recognition,” \textit{Neural Comput.}, vol. 7, no. 6, pp. 1289–1303, 1995.

\bibitem{b28} N. Arica and F. T. Yarman-Vural, “Optical Character Recognition for Cursive Handwriting,” \textit{IEEE Trans. Pattern Anal. Mach. Intell.}, vol. 24, no. 6, pp. 801–813, 2002.

\bibitem{b29} D. Peng, L. Jin, W. Ma \textit{et al.}, “A Fast and Accurate Fully Convolutional Network for End-to-End Handwritten Chinese Text Segmentation and Recognition,” in \textit{Proc. Int. Conf. on Document Analysis and Recognition (ICDAR)}, 2019, pp. 25–30.

\bibitem{b30} D. Peng, L. Jin, Y. Wu, Z. Wang, and M. Cai, “Recognition of Handwritten Chinese Text by Segmentation: A Segment-Annotation-Free Approach,” \textit{IEEE Trans. Multimedia}, vol. 23, pp. 3496–3507, 2020.

\bibitem{b31} D. Yu, X. Li, C. Zhang \textit{et al.}, “Towards Accurate Scene Text Recognition with Semantic Reasoning Networks,” \textit{IEEE Trans. Pattern Anal. Mach. Intell.}, vol. 44, no. 10, pp. 6594–6607, 2021.

\bibitem{b32} H. Zhang, F. Li, S. Liu \textit{et al.}, “DINO: DETR with Improved Denoising Anchor Boxes for End-to-End Object Detection,” in \textit{Proc. Int. Conf. on Learning Representations (ICLR)}, 2023.

\bibitem{b33} D. Coquenet, C. Chatelain, and T. Paquet, “DAN: A Segmentation-Free Document Attention Network for Handwritten Document Recognition,” \textit{IEEE Trans. Pattern Anal. Mach. Intell.}, vol. 45, no. 7, pp. 8227–8243, 2023.

\bibitem{b34} D. Coquenet, C. Chatelain, and T. Paquet, “Faster DAN: Multi-Target Queries with Document Positional Encoding for End-to-End Handwritten Document Recognition,” in \textit{Proc. Int. Conf. on Document Analysis and Recognition (ICDAR)}, 2023, pp. 182–199.

\bibitem{b35} T.-Y. Lin, P. Goyal, R. Girshick, K. He, and P. Doll{\'a}r, “Focal Loss for Dense Object Detection,” in \textit{Proc. IEEE Int. Conf. Computer Vision (ICCV)}, 2017, pp. 2980–2988.

\bibitem{b36} H. Rezatofighi, N. Tsoi, J. Gwak, A. Sadeghian, I. Reid, and S. Savarese, “Generalized Intersection over Union: A Metric and a Loss for Bounding Box Regression,” in \textit{Proc. IEEE/CVF Conf. Computer Vision and Pattern Recognition (CVPR)}, 2019, pp. 658–666.

\bibitem{b37} D. P. Kingma and J. Ba, “Adam: A Method for Stochastic Optimization,” in \textit{Proc. Int. Conf. on Learning Representations (ICLR)}, 2015.

\bibitem{b38} A. Graves, S. Fern{\'a}ndez, F. Gomez, and J. Schmidhuber, “Connectionist Temporal Classification: Labelling Unsegmented Sequence Data with Recurrent Neural Networks,” in \textit{Proc. 23rd Int. Conf. on Machine Learning (ICML)}, 2006, pp. 369–376.

\bibitem{b39} P. Krishnan and C. V. Jawahar, “HWNet v2.0: An Efficient Word Image Representation for Handwritten Documents,” \textit{Int. J. Document Anal. Recognit.}, vol. 19, pp. 167–177, 2016.

\bibitem{b40} B. Barz and J. Denzler, “The GoodNotes Handwriting Dataset,” 2020. [Online]. Available: \url{https://www.goodnotes.com/gnhk}

\bibitem{b41} A. Radford, J. Wu, R. Child, D. Luan, D. Amodei, and I. Sutskever, “Language Models are Unsupervised Multitask Learners,” \textit{OpenAI Blog}, vol. 1, no. 8, p. 9, 2019.

\bibitem{b42} S. M. Pizer, E. P. Amburn, J. D. Austin \textit{et al.}, “Adaptive Histogram Equalization and Its Variations,” \textit{Comput. Vision Graph. Image Process.}, vol. 39, no. 3, pp. 355–368, 1987.

\bibitem{b43} K. Heafield, “KenLM: Faster and Smaller Language Model Queries,” in \textit{Proc. Sixth Workshop on Statistical Machine Translation}, 2011, pp. 187–197.

\bibitem{b44} K. Song, X. Tan, T. Qin, J. Lu, and T.-Y. Liu, “MASS: Masked Sequence to Sequence Pre-training for Language Generation,” in \textit{Proc. Int. Conf. on Machine Learning (ICML)}, 2019.

\bibitem{b45} F. Kleber, S. Fiel, M. Diem, and R. Sablatnig, "CVL-Database: An Off-Line Database for Writer Retrieval, Writer Identification and Word Spotting," in \textit{Proc. Int. Conf. on Document Analysis and Recognition (ICDAR)}, 2013, pp. 560--564.

\bibitem{b46} Library of Congress, "George Washington Papers, Series 2, Letterbooks 1754-1799," [Online]. Available: \url{https://www.loc.gov/collections/george-washington-papers/}

\end{thebibliography}
\end{document}